\documentclass[runningheads]{llncs}
\usepackage[table]{xcolor}

\usepackage[T1]{fontenc}
\usepackage{graphicx,verbatim}
\usepackage[hidelinks,colorlinks=true,linkcolor=blue,citecolor=blue]{hyperref}
\usepackage{float}
\usepackage{pdfpages}
\usepackage{lipsum}
\usepackage{duckuments}
\usepackage{graphicx}
\usepackage{subcaption}
\usepackage{booktabs}
\usepackage[table]{xcolor}
\usepackage{amssymb}
\usepackage{multirow}
\usepackage{comment}
\usepackage{cite}
\usepackage{hyperref}
\usepackage{amsmath}

\newcommand{\res}[2]{#1 \scriptsize{$\pm #2$}}

\begin{document}
\title{ULF-Synth: Physics-Guided Ultra-Low-Field MRI Enhancement for Pediatric Neuroimaging}

% \title{ULF-Synth: Physics-Informed Pediatric Ultra Low-Field MRI Generative Enhancement}

\titlerunning{ULF-Synth: Physics-Guided Ultra Low-Field MRI Enhancement}
% \begin{comment}  %% Removed for anonymized MICCAI submission

\author{
Toufiq Musah\inst{1}\and
Salvatore Calcagno\inst{2} \and
Federica Proietto Salanitri\inst{2} \and
Xiaomeng Li\inst{4} \and
Maruf Adewole\inst{3} \and
Marawan Elbatel\inst{4}
}

\authorrunning{Toufiq Musah et al.}
\institute{
Kwame Nkrumah University of Science and Technology, Kumasi, Ghana \email{tmusah@st.knust.edu.gh}
\and
University of Catania, Catania, Italy
\and
Medical Artificial Intelligence Lab, Lagos, Nigeria
\and
The Hong Kong University of Science and Technology, Hong Kong SAR, China
}
  
% \end{comment}

\maketitle              % typeset the header of the contribution

\raggedbottom

\begin{abstract}
%\lipsum[1]

Ultra-low-field (ULF) MRI offers portable and accessible neuroimaging but suffers from reduced signal-to-noise ratio and limited spatial resolution compared to high-field (HF) systems. Acquiring paired ULF–HF data for supervised enhancement is often difficult, particularly in resource-limited settings. We introduce ULF-Synth, a framework that combines: (i) acquisition-based synthesis of realistic ULF images from HF volumes to create large-scale paired training data, (ii) a spatial-frequency domain objective that prioritizes recovery of high-frequency anatomical detail. This formulation is architecture-agnostic, consistently improving structural similarity and perceptual fidelity across encoder–decoder, adversarial, and diffusion-based translation models. When trained exclusively on synthetic data, the resulting models generalize effectively to real 64\,mT ULF acquisitions, improving downstream multiclass brain segmentation and achieving higher radiologist preference and diagnostic acceptability in a blinded reader study. These findings demonstrate that synthetic paired supervision provides a practical and scalable pathway for enhancing ULF MRI without requiring real paired acquisitions. Code, Models \& Dataset is available here: \url{https://github.com/toufiqmusah/ULF-Synth}


\end{abstract}

\section{Introduction}
Ultra-low-field (ULF) magnetic resonance imaging (MRI) systems (0.01–0.1T) offer portable, low-cost neuroimaging without the infrastructure demands of conventional 1.5T–3T scanners \cite{ULF-Systems}. Their accessibility has enabled deployment in medically underserved regions and bedside clinical environments, including pediatric and critical care settings \cite{UNITY, Brain-ULF-2}. However, reduced field strength limits signal-to-noise ratio, spatial resolution, and field homogeneity, resulting in degraded anatomical detail and increased artifacts \cite{Limit-ULF}. These limitations can constrain diagnostic confidence, necessitating validation with high-field MRI and or other imaging modalities where available \cite{Brain-ULF-2, Limit-ULF-2}. Therefore, improving image quality in ULF MRI is essential to improve its standalone clinical utility.

% Brain-ULF

Image enhancement and super-resolution methods have been explored to mitigate ULF quality limitations. Several approaches make use of paired high-field and ultra-low-field acquisitions to learn cross-domain translation \cite{LoHiResGAN,SFNet,GAMBAS,MRIQT}, while others adapt general MRI super-resolution architectures to the ULF setting \cite{SynthSR-1, SynthSR-2}. More recently, implicit neural representation (INR)-based approaches have explored subject-specific and unpaired enhancement \cite{ULF-INR}. Although effective, paired translation methods typically rely on scarce paired datasets acquired at specific field strengths (3T to 0.064T). Additionally, most enhancement frameworks optimize pixel-wise or perceptual objectives without explicitly enforcing MRI physics consistency across spatial-frequency domains. As a result, performance may depend on the similarity between training and deployment settings.

To address these challenges, we propose ULF-Synth, an MRI physics-guided ULF-to-HF synthesis framework that constructs paired supervision through a physically grounded degradation process, and enforces multi-domain consistency during enhancement. First, we introduce an MRI physics-based simulation pipeline that synthesizes ULF characteristics in HF scans, enabling scalable paired training. Second, we propose a composite loss integrating spatial-domain L1 supervision, frequency-aware weighted k-space consistency, and gradient regularization to preserve fine anatomical structure. Importantly, this supervision is model-agnostic and can be integrated into diverse image-to-image translation methods. Third, we evaluate the framework across adversarial, diffusion, and encoder–decoder architectures. Image enhancement performance is quantified on the synthetic ULF-to-HF dataset, and further on real out-of-distribution ULF scans using downstream multiclass segmentation of hippocampal and basal ganglia structures. A blinded expert reader study ranks our best-performing method against state-of-the-art baselines and evaluates clinical acceptability based on anatomical fidelity and overall diagnostic quality.

\section{Method}
\begin{figure}[t]
    \centering
    \includegraphics[width=1\textwidth]{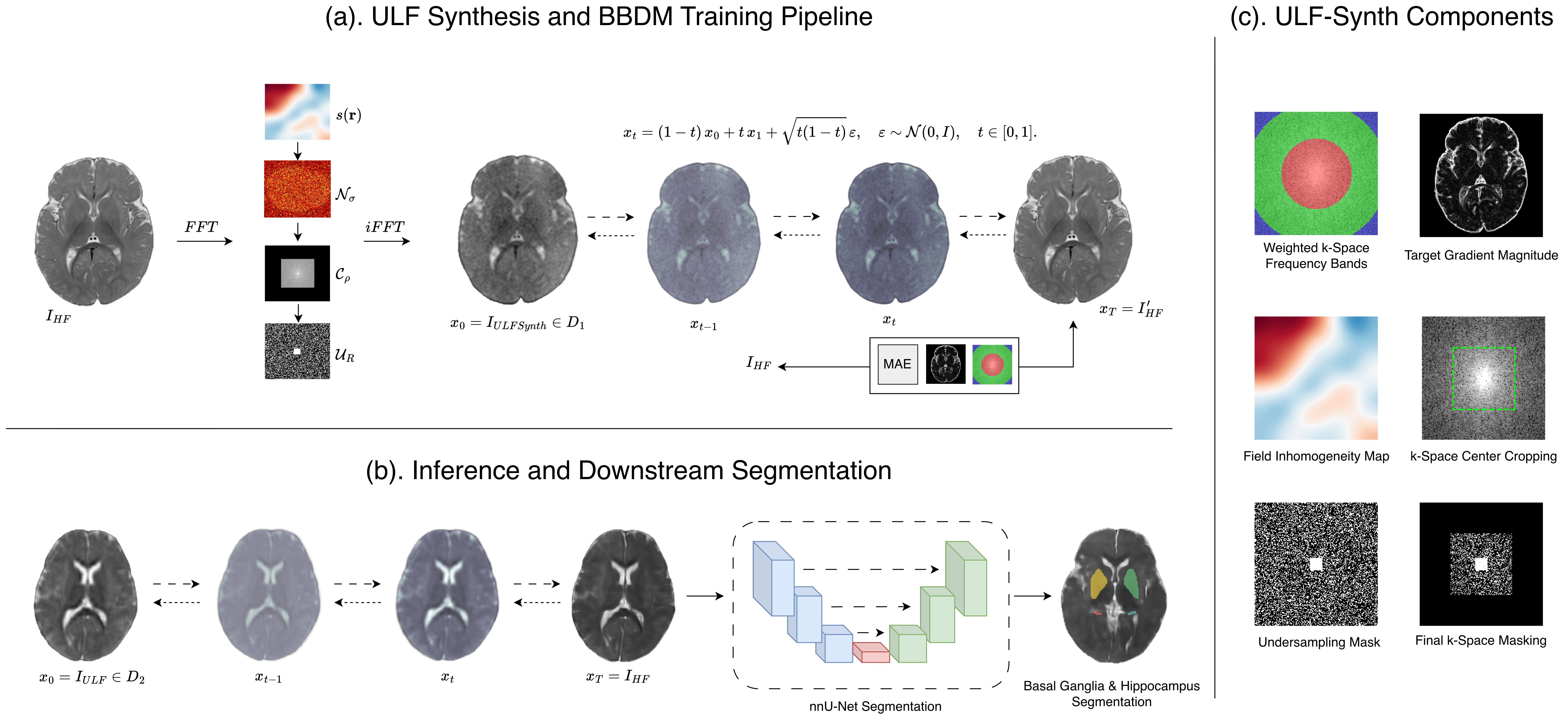}
    \caption{\textbf{ULF-Synth framework overview.} (\textbf{a}). HF volumes are degraded via sequential image- and $k$-space operations to generate synthetic ULF pairs, used in training a model with the composite loss. (\textbf{b}). Real ULF scans are enhanced and passed to a segmentation model for downstream evaluation. (\textbf{c}). Various components of the framework visualized.}
    \label{fig:main_image}
    \vspace{-0.5cm}
\end{figure}

\subsection{Dataset}
We curated a pediatric 1.5T T2-weighted MRI corpus of 833 publicly available volumes \cite{Dataset-D1}. The dataset was partitioned into 666 training, 83 validation, and 84 test volumes. Synthetic ULF counterparts were generated for each HF scan, yielding the paired synthetic training corpus denoted as $\mathcal{D}_1 = \{(x_i^{\text{ULF}}, x_i^{\text{HF}})\}$ . To assess out-of-distribution generalization we used 79 publicly available multisite real ULF brain MRI volumes acquired on a 64\,mT Hyperfine scanners, $\mathcal{D}_2$ \cite{LISA-D2}. % These volumes were split into 64 training and 15 testing samples for a downstream segmentation task and a radiologist reader study.

\subsection{Ultra-Low-Field MRI Synthesis}

% Exisiting work what does it do, no paired data, to this end we propose an ultra low field MRI synthesis strategy. specifically, to generate...etc

A major challenge in ULF enhancement is the scarcity of paired HF–ULF acquisitions \cite{LoHiResGAN}. Most prior approaches rely on co-registered data collected at specific field strengths, limiting scalability and generalization. While methods such as MRIQT\cite{MRIQT} synthesize ULF volumes, they require paired HF–ULF data to train the underlying translation model. To generate paired training data without requiring HF–ULF acquisitions, we simulate ULF measurements from HF volumes using an MRI physics-based degradation process. The pipeline models the dominant effects that distinguish ULF imaging (0.064\,T) from conventional 1.5\,T MRI, including reduced signal, field inhomogeneity, relaxation-induced decay, bandwidth limitation, undersampling, and noise. All operations are applied sequentially, and parameters are sampled per volume to reflect the variability of real portable scanners.

\paragraph{Image-space degradation.}
Let $x^{\text{HF}}$ be a 3D HF magnitude volume. We first apply a spatially varying bias field characteristic of low-field permanent-magnet systems and relaxation effects in image space to obtain a physically degraded complex volume $\tilde{x}$:

\begin{equation}
    \tilde{x}(\mathbf{r}) =
        s(\mathbf{r})\,
        x^{\text{HF}}(\mathbf{r})\,
        e^{-\mathrm{TE}/T_2^{*}(\mathbf{r})}\,
        e^{j\phi(\mathbf{r})},
    \label{eq:image_space}
\end{equation}

where $s(\mathbf{r})$ is a smooth ellipsoidal coil sensitivity map that models the spatially varying response of a portable permanent-magnet geometry \cite{Coil-Sensitivity, Coil-Sensitivity-2}, falling off gradually from the centre with a minimum of 30\% sensitivity at the edges. The exponential attenuation term models $T_2^*$ related signal decay, where the effective relaxation rate is determined by the intrinsic tissue $T_2$ and spatial gradients of a smooth random $B_0$ inhomogeneity field $B_0(\mathbf{r})$:

\begin{equation}
    \frac{1}{T_2^{*}(\mathbf{r})} = \frac{1}{T_2} + k\,\|\nabla B_0(\mathbf{r})\|.
    \label{eq:t2star}
\end{equation}

The phase term $e^{j\phi(\mathbf{r})}$, also derived from $B_0(\mathbf{r})$, simulates spatially varying dephasing. Parameters are sampled as $T_2 \in [0.06,\,0.10]$\,s, $\mathrm{TE} \in [0.08,\,0.15]$\,s, and $B_0$ strength $\in [0.02,\,0.05]$.

\paragraph{$k$-Space Degradation and Reconstruction.}
The degraded image $\tilde{x}$ is then transformed to $k$-space and subject to acquisition-level corruptions \cite{k-Space}:

\begin{equation}
    x^{\text{ULF}} =
        \left|\,\mathcal{F}^{-1}\!\left(
            \mathcal{U}_R\,\mathcal{C}_\rho\,\mathcal{N}_\sigma\,
            \mathcal{F}\{\tilde{x}\}
        \right)\right|.
    \label{eq:kspace}
\end{equation}

$\mathcal{N}_\sigma$ adds complex Gaussian noise in $k$-space, producing Rician-distributed magnitudes after reconstruction; noise variance $\sigma^2$ is scaled relative to the object's mean signal power to simulate target SNR conditions. 

$\mathcal{C}_\rho$ models the limited gradient bandwidth of ULF systems by retaining only a central fraction $\rho \in [0.45,\,0.55]$ of $k$-space and zero-padding to the original size, attenuating high-frequency anatomical detail. 

Finally, $\mathcal{U}_R$ applies structured undersampling with acceleration $R \in \{2,3\}$, with a fully sampled central region (fraction $\in [0.20,\,0.30]$) to preserve low-frequency content. All parameters are drawn independently per volume from uniform distributions over the ranges stated, and the pipeline is fully deterministic given a fixed random seed, enabling reproducible construction of the paired training corpus $\mathcal{D}_1$. The degradation parameters and pipeline design were empirically tuned through iterative qualitative assessment and radial power spectrum analysis to match the characteristics of real 0.064\,T acquisitions.

% title should highlight the name of your component or novelty (Physics Informed Frequency Based Optimization),
\subsection{Structured Spatial–Frequency Domain Supervision}

We formulate ULF enhancement as a supervised regression problem with structured constraints in both spatial and frequency domains, jointly enforcing voxel-level fidelity, edge preservation, and band-aware spectral consistency to promote anatomically coherent and acquisition-consistent reconstruction.

% We frame ULF enhancement as a supervised regression problem and optimize a composite loss that operates in both spatial and frequency domains, designed to preserve anatomical fidelity, recover fine structural details, and maintain consistency with the underlying acquisition physics:

\begin{equation}
    \mathcal{L}_{\text{total}} = \lambda_{\text{img}}\,\mathcal{L}_{\text{img}}
    + \lambda_{k}\,\mathcal{L}_{k}
    + \lambda_{\nabla}\,\mathcal{L}_{\nabla},
    \label{eq:total_loss}
\end{equation}

% \paragraph{Voxel-Level Fidelity Constraint.}
A voxel-wise $\ell_1$ loss establishes baseline correspondence between the generated volume $\hat{x}$ and the high-field reference $x$:

\begin{equation}
    \mathcal{L}_{\text{img}} = \|\hat{x} - x\|_1.
\end{equation}

\paragraph{Band-Weighted Log-Spectrum Consistency.}
Direct supervision in $k$-space encourages the model to reproduce the correct frequency content, which is especially important for high-frequency structures such as tissue boundaries. We decompose the Fourier magnitude spectrum into three radial bands (low, mid, and high) based on normalized distance from the $k$-space centre. For each band $\ell \in \{\text{low},\text{mid},\text{high}\}$ we compute the $\ell_1$ loss on the log-transformed magnitudes and combine them with band-specific weights:

\begin{equation}
    \mathcal{L}_{k} = \frac{
        \displaystyle\sum_{\ell} w_\ell\,
        \left\|
            M_\ell \odot
            \Bigl(
                \log(1 + |\mathcal{F}\{\hat{x}\}|)
                - \log(1 + |\mathcal{F}\{x\}|)
            \Bigr)
        \right\|_1
    }{
        \displaystyle\sum_{\ell} w_\ell
    },
    \label{eq:kspace_loss}
\end{equation}

where $\mathcal{F}$ denotes the centered 3D Fourier transform and $M_\ell$ is a binary mask selecting coefficients in band $\ell$. Since the available HF references are magnitude reconstructions without reliable phase information, supervision is applied to the log-magnitude spectrum. The log transformation amplifies weak high-frequency components that would otherwise be dominated by low-frequency energy. We empirically set band weights $\mathbf{w} = [1.5, 1.0, 2.0]$ to emphasize high-frequency recovery while maintaining global intensity consistency.

\paragraph{Structural Gradient Regularization.}
To sharpen edges and preserve brain folds, we penalize differences in spatial image gradients, approximated by finite differences along each axis:
\begin{equation}
    \mathcal{L}_{\nabla} = \bigl\|\,|\nabla\hat{x}| - |\nabla x|\,\bigr\|_1,
\end{equation}
where $|\nabla\,\cdot\,|$ denotes the Euclidean norm of the three
directional finite differences.

\section{Experimental Results}
% Experiment section.
% One Small paragraph for 
% Public Dataset.
% Baselines.
% Implementation Details.
% Metrics (if simple like F-1 score, do not describe)

% Quantitative Analysis (Refer to Table 1,  and include relative numbers in text (e.g.,g 5% improvement compared to x)
% one Table for generation, and segmentation
% Table 1 show our results. relative numbers and percentages in text. why?

% Qualitative results 
% Ablation Study (Table here with checkmarks of components)
% Adding your component, how does it increase the performance?

% \subsection{Dataset}
% \lipsum[1]

% \lipsum[2]
\noindent{\textbf{Baselines.}} The proposed ULF-Synth paradigm comprising synthetic ULF–HF paired data ($\mathcal{D}_1$) and spatial-frequency constrained supervision, is benchmarked across three representative image-to-image translation methods: a U-Net \cite{UNet} based translation method, nnU-Net translation \cite{nnUNet-Translation} (nnU-Net${^{T}}$); a conditional generative adversarial network \cite{GANs}, Pix2Pix \cite{Pix2Pix}; and a probabilistic diffusion-based approach \cite{DDPM}, Brownian-Bridge Diffusion Model \cite{BBDM-1, BBDM-2}. All models are trained on $\mathcal{D}_1$ under identical settings for approximately 40 hours.

%\noindent\textbf{{Ablations.}}
%Ablation experiments use the nnU-Net translation backbone trained for 250 epochs and progressively add loss components: (i) $\ell_1$ baseline only, (ii) $\ell_1 + \mathcal{L}_\nabla$, (iii) $\ell_1 +\mathcal{L}_k$ with uniform band weights, and (iv) the full composite objective with weighted bands. 

\noindent{\textbf{Downstream Segmentation.}} 
To assess clinical utility, we measure the effect of enhancement as a preprocessing stage for multiclass segmentation of the left and right hippocampus and basal ganglia (caudate and lentiform nuclei) on real ULF volumes acquired on a 64\,mT Hyperfine scanner \cite{LISA-D2}, $\mathcal{D}_2$. For each enhancement method, $\mathcal{D}_2$ is enhanced and used to train a default nnU-Net segmentation model \cite{nnU-Net}, for 250 epochs in a 5-fold cross-validation configuration.

% The resulting segmentation model is then evaluated on the correspondingly enhanced held-out test set. We report multiclass Dice coefficient, 95th-percentile Hausdorff distance (HD95), average symmetric surface distance (ASSD), and relative volume error (RVE).

\noindent{\textbf{Reader study.}}
We complement quantitative evaluation with a blinded radiologist preference study in which enhanced volumes are ranked against the unprocessed ULF baseline in randomized order. 

\noindent{\textbf{Evaluation Metrics.}}
Synthesis quality is assessed using Structural Similarity Index (SSIM), Multi-Scale SSIM (MS-SSIM), Learned Perceptual Image Patch Similarity (LPIPS), and Peak Signal-to-Noise Ratio (PSNR). Segmentation performance is evaluated using the Dice Similarity Coefficient (DSC), 95th percentile Hausdorff Distance (HD95), Average Symmetric Surface Distance (ASSD), and Relative Volume Error (RVE).

\begin{figure}[t]
    \centering
    \includegraphics[width=1\textwidth]{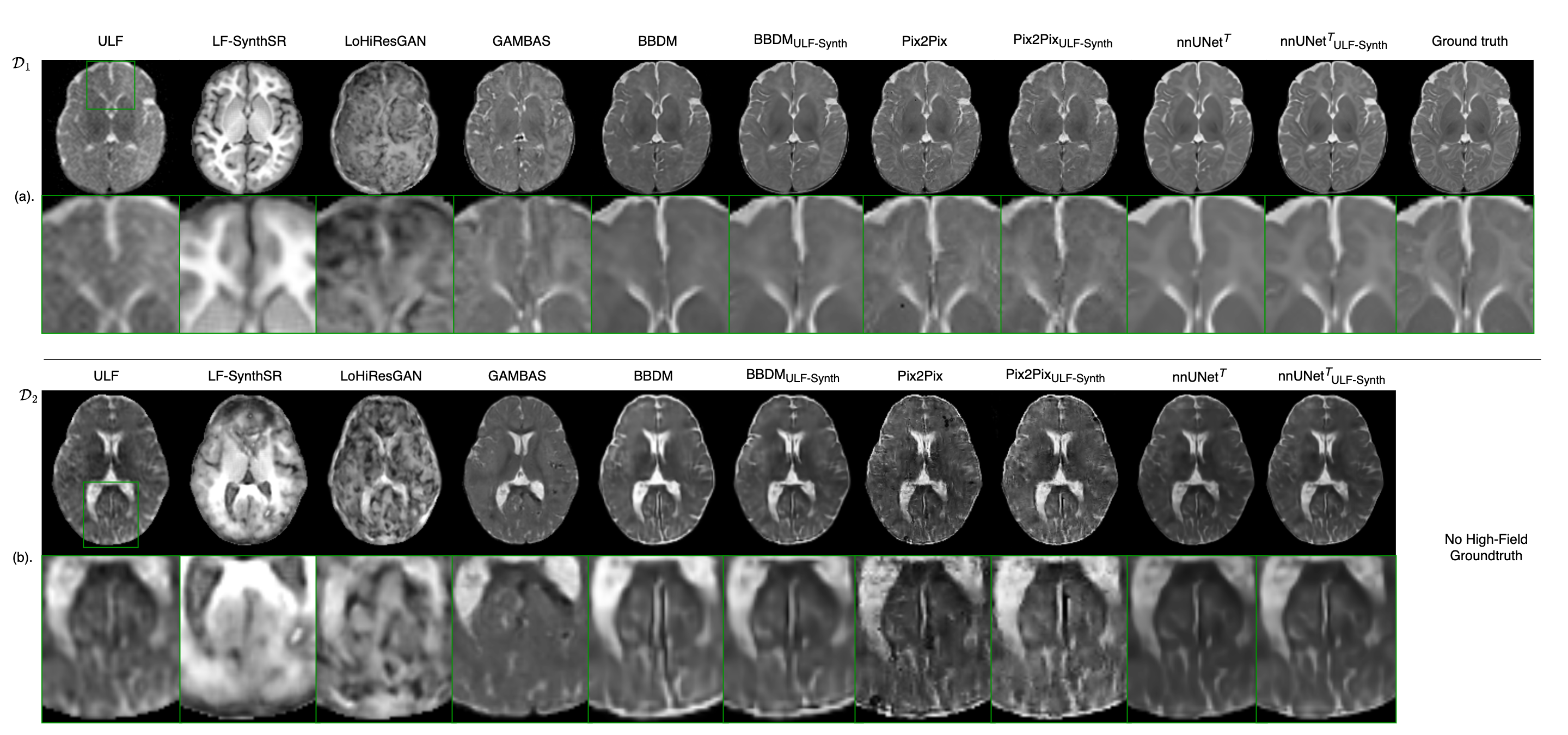}
    \caption{\textbf{Qualitative comparison of ULF image enhancement.}
    \textbf{(a)} Results on synthetic $\mathcal{D}1$.
    \textbf{(b)} Results on real clinical ULF acquisitions ($\mathcal{D}_2$). ULF-Synth models generalize effectively to out-of-distribution data.}
    \label{fig:qualitative_results}
    \vspace{-0.5cm}
\end{figure}

\subsection{Super-Resolution Performance on Synthetic Pairs ($\mathcal{D}_1$)}

Table~\ref{tab:sr_results} and Fig \ref{fig:qualitative_results} (a) show quantitative and qualitative synthesis performance on $\mathcal{D}_1$ respectively. Off-the-shelf state-of-the-art methods exhibit limited structural fidelity, with SSIM values below 0.75 and PSNR below 21 dB, reflecting a domain gap between conventional enhancement methods and the given ULF enhancement setting, rather than intrinsic methodological limitations. When trained directly on $\mathcal{D}_1$, the baselines (BBDM\cite{BBDM-1, BBDM-2}, Pix2Pix\cite{Pix2Pix}, nnU-Net${^{T}}$ \cite{nnUNet-Translation}) outperform prior methods. Among the baselines, nnU-Net${^{T}}$ achieves the best results (SSIM 0.9458, PSNR 28.69 dB). nnU-Net${^{T}_{ULF-Synth}}$ achieves the best overall performance (where the subscript $_{ULF-Synth}$ denotes training on $\mathcal{D}_1$ with the spatial-frequency domain objective). We therefore refer to nnU-Net${^{T}_{{ULF-Synth}}}$ as ULF-Synth.

% We emphasize that the spatial-frequency components ($\mathcal{L}_k$, $\mathcal{L}_\nabla$) are evaluated exclusively on the enhancement task (Table 1). For downstream segmentation, we employ only the baseline enhancement models trained on synthetic data to isolate the impact of synthesis quality.

% (SSIM 0.9466, MS-SSIM 0.9864, PSNR 28.76 dB).

\begin{table*}[ht]
\centering
\caption{\textbf{Super-Resolution Performance on $\mathcal{D}_1$}. Off-the-shelf methods ($^\dagger$) are applied without task-specific retraining. \textbf{Bold} indicates best overall performance; \underline{underline} indicates second-best.}
\label{tab:sr_results}
\rowcolors{2}{gray!10}{white}
\setlength{\tabcolsep}{3pt}
\renewcommand{\arraystretch}{1.2}
\begin{tabular}{lcccc}
\toprule
\textbf{Method}
    & \textbf{SSIM} ($\uparrow$)
    & \textbf{MS-SSIM} ($\uparrow$)
    & \textbf{LPIPS} ($\downarrow$)
    & \textbf{PSNR} ($\uparrow$) \\
\midrule
LF-SynthSR$^\dagger$~\cite{SynthSR-1, SynthSR-2}
    & \res{0.5741}{0.1642}
    & \res{0.4343}{0.1491}
    & \res{0.2268}{0.0443}
    & \res{13.97}{1.97}  \\
LoHiResGAN$^\dagger$~\cite{LoHiResGAN}
    & \res{0.7109}{0.0983}
    & \res{0.7629}{0.0707}
    & \res{0.1732}{0.0362}
    & \res{18.44}{2.24}  \\
GAMBAS$^\dagger$~\cite{GAMBAS}
    & \res{0.7438}{0.0849}
    & \res{0.8418}{0.0452}
    & \res{0.1257}{0.0252}
    & \res{20.99}{1.90}  \\
\midrule
BBDM~\cite{BBDM-1, BBDM-2}
    & \res{0.8833}{0.0513}
    & \res{0.9372}{0.0314}
    & \res{0.0420}{0.0136}
    & \res{22.41}{2.77}  \\
    
Pix2Pix~\cite{Pix2Pix}
    & \res{0.8943}{0.0459}
    & \res{0.9539}{0.0257}
    & \res{\textbf{0.0395}}{0.0133}
    & \res{23.60}{2.99}  \\
nnU-Net${^{T}}$ ~\cite{nnUNet-Translation}
    & \underline{\res{0.9458}{0.0203}}
    & \underline{\res{0.9853}{0.0066}}
    & \res{0.0480}{0.0155}
    & \underline{\res{28.69}{3.01}}  \\
\midrule
    
BBDM$_{ULF-Synth}$
    & \res{0.8986}{0.0477}
    & \res{0.9441}{0.0316}
    & \res{0.0580}{0.0210}
    & \res{22.92}{3.08}  \\
    
Pix2Pix$_{ULF-Synth}$
    & \res{0.9123}{0.0482}
    & \res{0.9599}{0.0257}
    & \res{0.0420}{0.0136}
    & \res{24.37}{2.94}  \\
    
nnU-Net${^{T}_{ULF-Synth}}$
    & \textbf{\res{0.9466}{0.0207}}
    & \textbf{\res{0.9864}{0.0074}}
    & \underline{\res{0.0400}{0.0136}}
    & \textbf{\res{28.76}{2.91}}  \\
\bottomrule
\end{tabular}
\end{table*}

\subsection{Evaluation on Real ULF Data ($\mathcal{D}_2$)}

\paragraph{Downstream Segmentation.} To assess generalization, we evaluate enhancement methods as a preprocessing step for multiclass segmentation on real 64\,mT ULF volumes ($\mathcal{D}_2$) \cite{toufiq-ulf}. Table~\ref{tab:seg_results} reports quantitative segmentation performance, and Fig.~\ref{fig:qualitative_results} (b) illustrates qualitative enhancement results. nnU-Net${^{T}}_{ULF-Synth}$ trained on $\mathcal{D}_1$ with  the spatial-frequency constraints achieves the strongest overall segmentation performance, with consistent improvements over the ULF baseline across boundary-sensitive metrics. These results indicate that models trained on synthetic paired data ($\mathcal{D}_1$) generalize effectively to real ULF acquisitions. Statistical significance further assessed using the paired Wilcoxon signed-rank test, comparing against the ULF-baseline segmentation results. All models trained on $\mathcal{D}_1$ showed statistically significant improvements over the baseline.

\begin{table*}[h!]
\centering
\caption{\textbf{Downstream Segmentation on Real ULF Data ($\mathcal{D}_2$).} \textbf{Bold} indicates best performance; \underline{underline} indicates second-best. *$p<0.05$ (paired Wilcoxon signed-rank vs. ULF-Baseline). Metrics are per-class averages over the six foreground liver segments; RVE is computed on the whole liver volume. \label{tab:seg_results}}
\rowcolors{2}{gray!10}{white}
\setlength{\tabcolsep}{3pt}
\renewcommand{\arraystretch}{1.2}
\begin{tabular}{lcccc}
\toprule
\textbf{Method} & \textbf{DSC} ($\uparrow$) & \textbf{HD95 (mm)} ($\downarrow$) & \textbf{ASSD} ($\downarrow$) & \textbf{RVE (\%)} ($\downarrow$) \\
\midrule
ULF-Baseline
    & \res{0.7106}{0.1609} & \res{4.76}{7.10} & \res{1.59}{2.98} & \res{2.97}{15.13} \\
LF-SynthSR~\cite{SynthSR-1, SynthSR-2}
    & \res{0.6514}{0.2250} & \res{6.58}{9.81} & \res{3.15}{6.67} & \res{2.53}{16.44} \\
LoHiResGAN~\cite{LoHiResGAN}
    & \res{0.7403}{0.0791} & \res{2.40}{0.75}\textsuperscript{$*$} & \res{0.75}{0.30} & \res{0.55}{14.36} \\
GAMBAS~\cite{GAMBAS}
    & \res{0.7514}{0.0821} & \res{2.37}{0.81}\textsuperscript{$*$} & \res{0.72}{0.31} & \textbf{\res{0.33}{12.94}}\textsuperscript{$*$} \\
\midrule
BBDM~\cite{BBDM-1}
    & \res{0.7589}{0.0821}\textsuperscript{$*$} & \res{2.28}{0.75}\textsuperscript{$*$} & \res{0.70}{0.31}\textsuperscript{$*$} & \res{0.98}{13.75}\textsuperscript{$*$} \\
Pix2Pix~\cite{Pix2Pix}
    & \res{0.7597}{0.0807}\textsuperscript{$*$} & \res{2.28}{0.77}\textsuperscript{$*$} & \res{0.69}{0.31}\textsuperscript{$*$} & \res{0.99}{13.51}\textsuperscript{$*$} \\
nnU-Net${^{T}}$~\cite{nnUNet-Translation}
    & \underline{\res{0.7619}{0.0808}}\textsuperscript{$*$} & \textbf{\res{2.25}{0.79}}\textsuperscript{$*$} & \textbf{\res{0.69}{0.32}}\textsuperscript{$*$} & \res{1.41}{13.71}\textsuperscript{$*$} \\
\midrule
BBDM$_{ULF-Synth}$
    & \res{0.7517}{0.1100}\textsuperscript{$*$} & \res{2.61}{3.40}\textsuperscript{$*$} & \res{0.87}{1.56}\textsuperscript{$*$} & \underline{\res{0.76}{13.03}}\textsuperscript{$*$} \\

Pix2Pix$_{ULF-Synth}$
    & \res{0.7566}{0.0904}\textsuperscript{$*$} & \res{2.64}{3.30}\textsuperscript{$*$} & \res{0.82}{1.21}\textsuperscript{$*$} & \res{1.89}{13.79}\textsuperscript{$*$} \\

nnU-Net${^{T}}_{ULF-Synth}$
    & \textbf{\res{0.7625}{0.0820}}\textsuperscript{$*$} & \underline{\res{2.27}{0.80}}\textsuperscript{$*$} & \textbf{\res{0.69}{0.32}}\textsuperscript{$*$} & \res{1.22}{13.84}\textsuperscript{$*$} \\
\bottomrule
\end{tabular}
\end{table*}

\paragraph{Reader Study.} We further assessed perceptual quality through a blinded reader study on five randomly selected $\mathcal{D}_2$ subjects. For each case, the original ULF volume and four synthetically enhanced outputs (SynthSR, LoHiResGAN, GAMBAS, and nnU-Net${^{T}}$) were presented in randomized order. Three radiologists ranked the four enhancement methods and indicated diagnostic acceptability. 

Inter-rater agreement was strong (Spearman’s $\rho=0.80$; Kendall’s $W=0.90$), indicating consistent relative preferences. Method rankings differed significantly (Friedman $\chi^2=20.39$, $p<0.001$). Post-hoc pairwise testing (Bonferroni-corrected $\alpha=0.0083$) showed that GAMBAS \cite{GAMBAS} and nnU-Net${^{T}}$ were both preferred over SynthSR and LoHiResGAN, while no significant difference was observed between the top two approaches. Diagnostic acceptability reflected the same trend: GAMBAS and nnU-Net${^{T}}$ were deemed acceptable in all evaluations, whereas SynthSR and LoHiResGAN demonstrated lower acceptability rates.

\subsection{Ablation Study}

Table~\ref{tab:ablation_results} quantifies the relative contribution of spatial and frequency-domain supervision using nnU-Net${^{T}}$ trained for 250 epochs on $\mathcal{D}_1$. Gradient regularization ($\nabla$) improves SSIM by 0.86\% and reduces LPIPS by 16.0\%, but reduces PSNR by 3.9\%. Frequency-domain supervision alone yields stronger structural gains: uniform band weighting increases SSIM by 1.25\% and MS-SSIM by 0.52\%, while weighted band emphasis achieves the largest standalone MS-SSIM improvement (+0.64\%), reflecting improved multi-scale structural consistency. Spatial and spectral constraints produces the most balanced performance, with the largest SSIM gain (+1.35\%), and perceptual improvement (LPIPS reduced by 20.7\%).

%, and the highest PSNR increase (+0.81\%). 

\begin{table*}[h]
\centering
\caption{\textbf{Ablation study of loss components on synthesis}. All models trained for 250 epochs on $\mathcal{D}_1$ with nnU-Net Translation in combination with L1.}
\label{tab:ablation_results}

\rowcolors{2}{gray!10}{white}
\setlength{\tabcolsep}{1.8pt}
\renewcommand{\arraystretch}{1.2}

\begin{tabular}{lcccc}
\toprule
\textbf{Configuration}
    & \textbf{SSIM} ($\uparrow$)
    & \textbf{MS-SSIM} ($\uparrow$)
    & \textbf{LPIPS} ($\downarrow$)
    & \textbf{PSNR (dB)} ($\uparrow$) \\
\midrule

L1 only
    & \res{0.9144}{0.028}
    & \res{0.9737}{0.006}
    & \res{0.0681}{0.018}
    & \res{27.02}{2.27} \\

$\nabla$
    & \res{0.9223}{0.029}
    & \res{0.9738}{0.011}
    & \res{0.0572}{0.015}
    & \res{25.98}{3.63} \\
    
\midrule
k$_{uniform}$
    & \res{0.9258}{0.029}
    & \res{0.9788}{0.009}
    & \res{0.0656}{0.018}
    & \res{27.09}{3.28} \\

k$_{weighted}$
    & \res{0.9237}{0.029}
    & \res{\textbf{0.9799}}{0.009}
    & \res{0.0647}{0.018}
    & \res{27.06}{3.10} \\

\midrule

k$_{uniform}$ + $\nabla$
    & \res{0.9265}{0.029}
    & \res{0.9770}{0.010}
    & \res{0.0555}{0.016}
    & \res{27.19}{3.83} \\

k$_{weighted}$ + $\nabla$
    & \res{\textbf{0.9267}}{0.028}
    & \res{0.9775}{0.008}
    & \res{\textbf{0.0540}}{0.015}
    & \res{\textbf{27.24}}{3.16} \\

\bottomrule
\end{tabular}

\end{table*}

\section{Conclusion}
We present ULF-Synth, a physics-guided framework for ULF MRI enhancement, combining synthetic paired supervision with structured spatial-frequency domain learning. By simulating acquisition-specific degradations directly from HF MRI, the proposed approach enables scalable construction of paired ULF--HF datasets. Under model-agnostic settings, ULF-Synth consistently improves structural fidelity and perceptual quality on synthetic data while generalizing effectively to real 64\,mT acquisitions. Downstream segmentation experiments and blinded radiologist evaluation further demonstrate that models trained exclusively on synthetic supervision can improve anatomical delineation and diagnostic acceptability on real ULF scans. Ablation studies indicate that synthetic paired supervision accounts for the largest performance gains, while the proposed multi-domain constraints provide complementary improvements in fine-detail preservation. These findings support physics-guided synthetic supervision as a practical and scalable pathway toward reliable ULF MRI enhancement in resource-constrained and portable neuroimaging settings.

\section{Limitations}
This work is limited to T2-Weighted imaging. Generalizability to other parameters such as T1 or flair remains an open question. While the effectiveness of transfer from synthetic to real ULF data is well-demonstrated, the fidelity of the underlying physical simulation model has not been independently validated against real ultra-low-field statistics. Further, although the spatial-frequency constraints provide qualitative improvements (see Fig. 2), their quantitative impact on synthesis metrics and downstream segmentation is moderate, suggesting that further hyperparameter tuning could lead to more pronounced benefits.

% - Limited to T2W. 
% - Simulation is shown to transfer, but not independently validated. 
% Improvements from frequency constrained losses, though good in qualitative from fig2, remain modest in both enhancement quality and downstream performance quantitatively. further hyperparameter tuning will be beneficial.

\section*{Acknowledgements}
This work was partially supported by the European Union -- Next Generation EU, Mission 4 Component 2 Line 1.3, through the PNRR MUR project PE0000013 -- FAIR ``Future Artificial Intelligence Research'' (CUP E63C22001940006). Marawan Elbatel is supported by the Hong Kong PhD Fellowship Scheme (HKPFS) from the Hong Kong Research Grants Council (RGC), and by the Belt and Road Initiative from the HKSAR Government.

\begin{credits}
% \subsubsection{\ackname} None.

\subsubsection{\discintname}

The authors have no competing interests to declare that are relevant to the content of this article.

\end{credits}

% \newpage

\bibliographystyle{splncs04}
\bibliography{biblio}

\end{document}